\documentclass[fleqn,11pt]{wlscirep}
\usepackage[utf8]{inputenc}
\usepackage[T1]{fontenc}

\usepackage{amsmath}

\usepackage{color}
\usepackage{xcolor}
\definecolor{rev1}{rgb}{0,0,0}
\definecolor{rev2}{rgb}{0,0,0}

\usepackage{dcolumn}% Align table columns on decimal point
\usepackage{bm}% bold math
\usepackage{subfigure}

\usepackage{soul}
\usepackage{amsmath,amssymb,amsfonts}
\usepackage{graphics}

\usepackage{algorithm}
\usepackage{algorithmic}
\usepackage{enumitem}
\setlist[itemize]{leftmargin=*}
\setlist[enumerate]{leftmargin=*}

\usepackage{comment}
\usepackage{hyperref}
\usepackage{lipsum}
\usepackage{tabularx}
\usepackage{multirow}
\usepackage{mathrsfs}
\usepackage{stackengine}

\title{Variational multiscale reinforcement learning for discovering reduced order closure models of nonlinear spatiotemporal transport systems}

\author[1*]{Omer San}
\author[1]{Suraj Pawar}
\author[2,3]{Adil Rasheed}

\affil[1]{School of Mechanical \& Aerospace Engineering, Oklahoma State University, Stillwater, OK 74078, USA.}
\affil[2]{Department of Engineering Cybernetics, Norwegian University of Science and Technology, 7465 Trondheim, Norway.}
\affil[3]{Department of Mathematics and Cybernetics, SINTEF Digital, 7034 Trondheim, Norway.}
\affil[*]{osan@okstate.edu}


\begin{abstract}

A central challenge in the computational modeling and simulation of a multitude of science applications is to achieve robust and accurate closures for their coarse-grained representations due to underlying highly nonlinear multiscale interactions. These closure models are common in many nonlinear spatiotemporal systems to account for losses due to reduced order representations, including many transport phenomena in fluids. Previous data-driven closure modeling efforts have mostly focused on supervised learning approaches using high fidelity simulation data. On the other hand, reinforcement learning (RL) is a powerful yet relatively uncharted method in spatiotemporally extended systems. In this study, we put forth a modular dynamic closure modeling and discovery framework to stabilize the Galerkin projection based reduced order models that may arise in many nonlinear spatiotemporal dynamical systems with quadratic nonlinearity. However, a key element in creating a robust RL agent is to introduce a feasible reward function, which can be constituted of any difference metrics between the RL model and high fidelity simulation data. First, we introduce a multi-modal RL (MMRL) to discover mode-dependant closure policies that utilize the high fidelity data in rewarding our RL agent. We then formulate a variational multiscale RL (VMRL) approach to discover closure models without requiring access to the high fidelity data in designing the reward function. Specifically, our chief innovation is to leverage variational multiscale formalism to quantify the difference between modal interactions in Galerkin systems. Our results in simulating the viscous Burgers equation indicate that the proposed VMRL method leads to robust and accurate closure parameterizations, and it may potentially be used to discover scale-aware closure models for complex dynamical systems.

\end{abstract}
\begin{document}

\flushbottom
\maketitle
\thispagestyle{empty}

% \noindent Please note: Abbreviations should be introduced at the first mention in the main text – no abbreviations lists. Suggested structure of main text (not enforced) is provided below.

\section*{Introduction}
% motivation, review, list applications

Reduced order models (ROMs) often refer to simplifications of high-fidelity models that capture dominant system dynamics using minimal computational resources. Over the past decades, we have witnessed an ever increasing number of reduced order modeling approaches and their enormous impact on fluid dynamics research \cite{lucia2004reduced,rowley2017model,taira2020modal,ahmed2021closures}. A chief motivation behind these approaches being introduced is to be able to use ROMs in multi-query applications such as control and optimization \cite{benner2015survey,peherstorfer2018survey}. 

Broadly speaking, closure modeling in fluid flow simulations refers to parameterizing the interactions between high-fidelity and coarse-grained descriptions. Although projection based ROMs have been utilized extensively in many fluid dynamics applications \cite{lucia2004reduced,rowley2017model,taira2020modal,ahmed2021closures}, they might yield inaccurate results when they are used in the under-resolved regime, i.e., when the number of modes is not large enough to capture the parameterized or transient dynamics of the underlying system \cite{snyder2022reduced}. Prior studies have suggested that closure models are efficacious in decreasing such modal truncation errors \cite{ahmed2021closures}. In fact, ROM closures can be viewed as correction or residual terms that are added to classical ROMs in order to model the effect of the discarded ROM modes in under-resolved simulations.

Consequently, an emerging thrust in modern ROM closure development efforts is to incorporate machine learning (ML) models \cite{milano2002neural,san2018neural,pawar2019deep,pan2018data,gupta2021neural,ahmed2021closures}. The last decade has seen the growth of data-driven modeling technologies (e.g., deep neural networks). So far, a substantial body of closure modeling works has focused on supervised learning \cite{san2018extreme,san2018neural,ahmed2020long}. A detailed discussion on these models can be found in a recent survey \cite{ahmed2021closures}. In principle, the problem of building a data-driven closure model from the multimodal datasets can be posed as an optimization and ML task. Although supervised learning techniques become more of commodity tools nowadays, reinforcement learning (RL) is relatively an uncharted approach in computational science communities. In contrast to many other data-driven supervised learning based approaches introduced for closures\cite{ahmed2021closures}, this powerful  approach can be formulated to tackle with the closure problem in an automated fashion. 

RL often provides a comprehensive iterative computational framework that implies modular goal-directed interactions of an agent with its environment. More recently, Novati et al. \cite{novati2021automating} demonstrated the power of RL in discovery of turbulence closure models in large-eddy simulation of turbulent flows. The RL framework being introduced in the turbulence modeling context is quite comprehensive, and might apply equally well to other coarse-grained reduced order modeling approaches. In relevant works \cite{benosman2020reinforcement,benosman2021reinforcement}, the feasibility of using RL to learn optimal ROM closures has been discussed. We also refer to a recent review \cite{garnier2021review} for the theory and application of RL approaches in fluid mechanics.

A fundamental question in RL is how to construct robust state, action and reward definitions relevant to the underlying problem. In this study, we put forth and examine the scale-aware RL mechanisms that automate closure modeling of projection based ROMs. Specifically, we first introduce a multi-modal RL (MMRL) approach, which discovers the mode dependant policies to stabilize the evolution of the truncated Galerkin system. One of the main objectives of our study is therefore to demonstrate how physical insights play a key role in designing an effective RL environment that converges to a robust control policy for learning and parameterizing the closure terms. 

In fact, a key element in forging a robust RL agent is to introduce an appropriate reward function, which can be, in principle, constituted of any difference metrics between the RL model and high fidelity simulation data. In this study, instead, we explore how RL can be formulated using a variational multiscale approach to discover closure models without requiring access to the high fidelity data in designing the reward function. The general concept of the variational multiscale framework has been first introduced in finite element community \cite{hughes1998variational,hughes2000large,hughes2001large,codina2018variational,john2016finite}, and its underpinning idea has been later adopted by ROM modellers \cite{stabile2019reduced,reyes2020projection,tello2020fluid,mou2021data,koc2021verifiability,ahmed2022physics}. In our study, we put forth a variational multiscale RL (VMRL) approach by leveraging this multiscale concept that introduces a natural hierarchy to quantify the difference between modal interactions in the Galerkin projection based ROM systems. Specifically, our work addresses the following questions: 
\begin{itemize} 

\item How can RL be used to discover reliable closure models in reduced order models of transport equations? 

\item Which parameterization processes lead to improved closure approaches that may reduce uncertainty in the evolution of projection coefficients of the Galerkin ROM systems?

\item How does a mode-dependent closure formulation affect overall predictive performance? 

\item What are the design considerations for formulating a feasible reward function that does not require access to the supervised training data? 

\end{itemize}
Therefore, the main goal of this paper is to address these questions in the context of closure model discovery for complex nonlinear spatiotemporal systems.

\section*{Methods}

\subsection*{Reduced order modeling}

\begin{figure}[htbp!]
\centering
\includegraphics[width=1.0\linewidth]{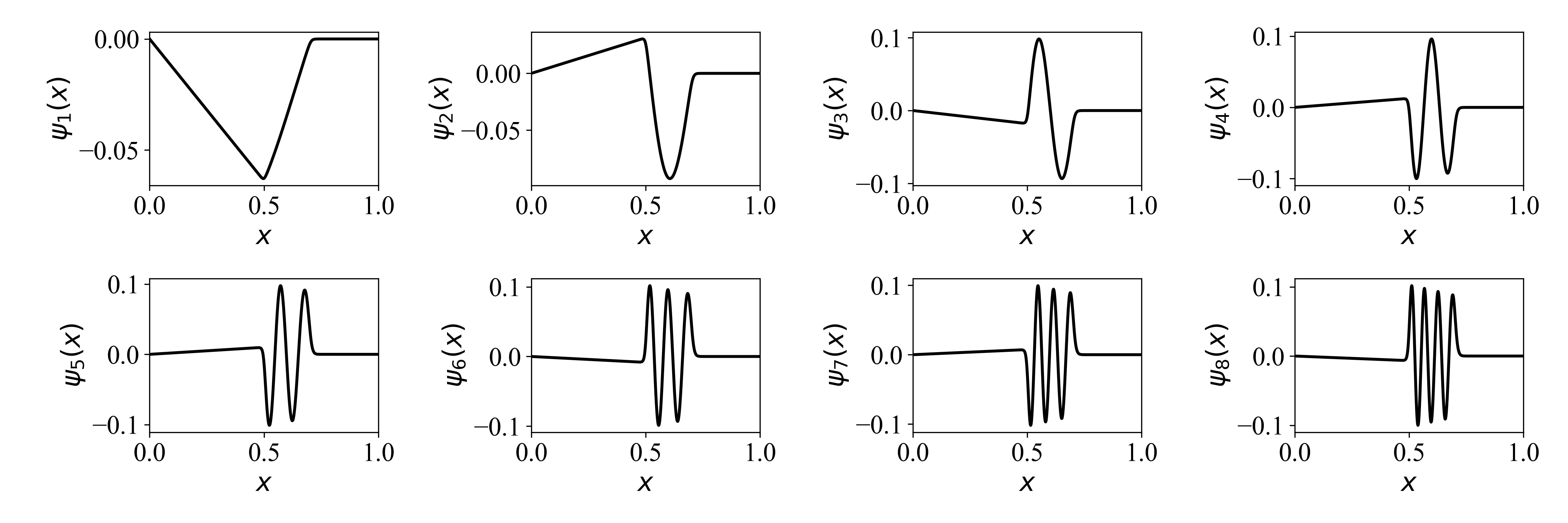}
\caption{Illustration of the first eight POD basis functions generated using a total of 500 data snapshots at $Re = 1000$.
}
\label{fig:basis}
\end{figure}

To illustrate the proposed approaches, we focus on the viscous Burgers equation, a generic partial differential equation that represents broad nonlinear transport phenomena in fluid dynamics, which is given as \cite{ahmed2021closures} 
\begin{align}\label{eq:burgers}
  \frac{\partial u}{\partial t} + u\frac{\partial u}{\partial x} = \nu \frac{\partial^2 u}{\partial x^2},
\end{align}
where $u$ refers to velocity, and $\nu$ is the kinematic viscosity (i.e., $\nu = 1/\mbox{Re}$ in dimensionless form, where Re refers to the Reynolds number). From a model reduction perspective, we highlight that Equation~(\ref{eq:burgers}) possesses the hallmarks of general nonlinear multidimensional advection-diffusion problems \cite{san2019artificial}. Defining our spatiotemporal domain, $x \in [0, 1]$ and $t \in [0, 1]$, the viscous Burgers equation admits an analytical solution in the form of
 \cite{san2019artificial,maleewong2011line}
\begin{align}
\label{eq:exact}
u(x,t) = \frac{\frac{x}{t+1}}{1+\sqrt{\frac{t+1}{t_0}}\exp( \frac{x^2}{4\nu(t+1)})},
\end{align}
where $t_0 = \exp(\frac{1}{8\nu})$. This closed form expression is used to generate snapshot data for our forthcoming model order reduction analysis. The database is constituted with Eq.~(\ref{eq:exact}) using $N = 1024$ spatial collocation points at each snapshot. We set $\nu=0.001$ (i.e., $Re =1000$) to generate our training data and our training database consists of 500 snapshots from $t=0$ to $t=1$. In considering such spatiotemporal system, we often use a modal decomposition to the field $u(x,t)$
\begin{align}\label{eq:base}
u(x,t) =\sum_{k=1}^{R} \alpha_k(t) \psi_k(x),
\end{align}
where $\alpha_k(t)$ and $\psi_k(x)$ refer to the $k$th modal coefficient and $k$th proper orthogonal decomposition (POD) basis function, respectively. Figure~\ref{fig:basis} shows the the first eight most energetic POD basis functions utilized in this study. We note that, without losing generality, one can also use Fourier harmonics \cite{san2018neural} or randomized orthogonal functions \cite{bistrian2022rod} to forge a set of basis functions.  

Once a set of spatial orthonormal modes (i.e., for $k=1,2, ..., R$) is obtained from snapshot data, we then apply the Galerkin projection (GP) to obtain a dynamical system that evolves in the latent space of $\alpha_k(t)$. The resulting ROM, denoted as GP in this study, becomes
\begin{equation} \label{eq:rom}
  \frac{d \alpha_k}{d t} =  \sum_{i=1}^{R} \mathfrak{L}^{i}_{k}\alpha_{i} + \sum_{i=1}^{R}\sum_{j=1}^{R} \mathfrak{N}^{ij}_{k}\alpha_{i}\alpha_{j}, \quad \forall \ k = 1, \ldots, R
\end{equation}
where
\begin{eqnarray}
  & & \mathfrak{L}^{i}_{k} = \bigg( \nu \frac{\partial^2 \psi_i(x) }{\partial x^2}, \psi_{k}(x) \bigg), \\
  & &  \mathfrak{N}^{ij}_{k} = \bigg( - \psi_i(x)\frac{\partial \psi_j(x)}{\partial x}, \psi_{k}(x) \bigg), \label{eq:roma7}
\end{eqnarray}
where the notion of $(\cdot, \cdot)$ represents the standard inner product. We note that these tensorial coefficients in GP model only depend on spatial modes, which are often precomputed from the available snapshot data when designing projection based ROMs. 

%Previous studies have shown that such GP based ROMs might yield inaccurate results, especially in convection dominated problems \cite{ahmed2021nonlinear}. Therefore, many efforts have been focused onto developing techniques to correct the evolution trajectory of GP models (e.g., see \cite{san2014proper} and references therein). In this study, we formulate this ROM closure problem using a deep RL modeling approach to provide a more general and modular framework for the closure model discovery. 

\subsection*{Closure modeling}

\begin{figure}[t]
\centering
\includegraphics[width=0.8\linewidth]{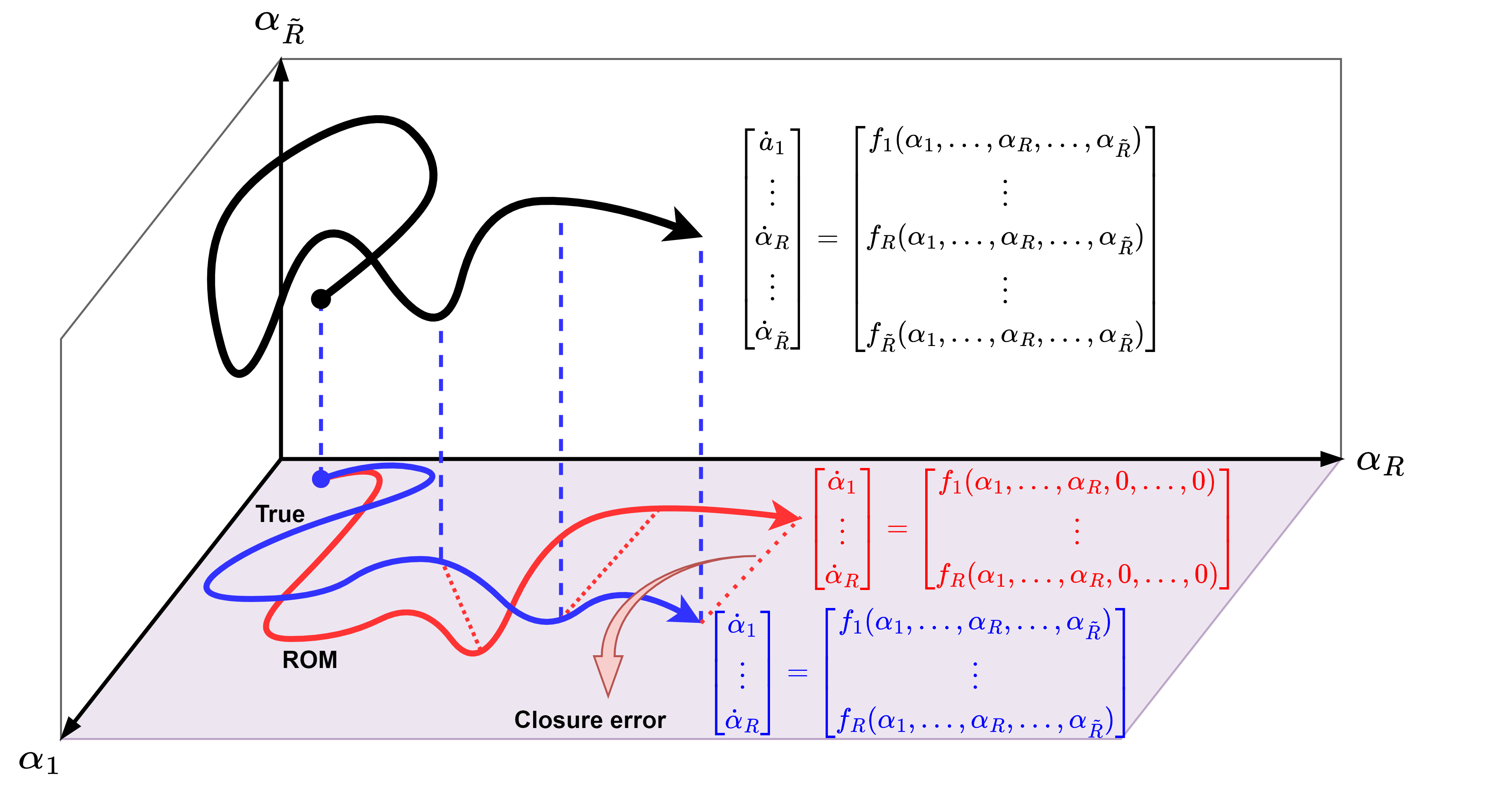}
\caption{A schematic overview of the closure modeling in a hypothetical three-dimensional latent space. When we truncate a model spanned in a higher dimensional state space (i.e.,  $\tilde{R}=3$ in the figure, a three-dimensional model spanned in $\alpha_1,\alpha_2,\alpha_3$) to a lower dimensional latent space (i.e., $R=2$ in the figure, a reduced order two-dimensional model spanned only in $\alpha_1,\alpha_2$), a closure error will be introduced due to the underlying nonlinear interactions. The main goal in closure modeling is to find a parameterized model, which is only function of resolved modal coefficients (i.e., $\alpha_1$, $\ldots$, $\alpha_R$) to minimize this closure error. Therefore, in this paper we formulate an RL problem to discover this closure model using the state variables (i.e., resolved modal coefficients). }
%$\mathbb{M} \in \mathbb{R}^{\tilde{R}}$ where
%$\mathbb{M} \in \mathbb{R}^{R}$ where
\label{fig:closure}
\end{figure}

We first illustrate the underlying closure modeling concept for a prototype demonstration as shown in Figure~\ref{fig:closure}. 
To formulate our ROM closure problem, we modify Eq.~\ref{eq:rom} by adding a functional form of distributed control. This control is often referred to as eddy viscosity approach that has strong roots in large eddy simulations to model or compensate the residual effects of the truncated scales \cite{borggaard2009bounded,akhtar2012new,san2014proper,san2015stabilized}. Therefore, the modified ROM becomes
\begin{equation} \label{eq:rom1}
  \frac{d \alpha_k}{d t} =  \sum_{i=1}^{R} \mathfrak{L}^{i}_{k}\alpha_{i} + \sum_{i=1}^{R} \tilde{\mathfrak{L}}^{i}_{k}\alpha_{i} + \sum_{i=1}^{R}\sum_{j=1}^{R} \mathfrak{N}^{ij}_{k}\alpha_{i}\alpha_{j}, \quad \forall \ k = 1, \ldots, R
\end{equation}
where the proposed closure term can be parameterized by defining an eddy viscosity coefficient $\eta$ as follows
\begin{equation}
\tilde{\mathfrak{L}}^{i}_{k} = \bigg(\eta \frac{\partial^2 \psi_i(x) }{\partial x^2}, \psi_{k}(x) \bigg).  \label{eq:closure1}
\end{equation}
Several techniques have been introduced to improve the accuracy of closure parameterizations, including definition of a nonlinear eddy viscosity model \cite{cordier2013identification} or dynamic closure models \cite{wang2012proper,rahman2019dynamic} that allow varying eddy viscosity in time (i.e., $\eta(t) \leftarrow \eta$). In this paper, we first formulate an RL environment and design an agent to discover this eddy viscosity parameter $\eta(t)$. We call this approach linear-mode RL (LMRL).

In their seminal work, {\"O}sth et al. \cite{osth2014need} further enhanced the closure theory emphasizing the modal eddy viscosity concept. The roots of such mode-dependent correction go back to the work of Rempfer and Fasel \cite{rempfer1993dynamics} in order to provide improved closure models.
These multi-modal closures can be specified as 
\begin{equation}
\tilde{\mathfrak{L}}^{i}_{k} =  \bigg(\eta_k \frac{\partial^2 \psi_i(x) }{\partial x^2}, \psi_{k}(x) \bigg).  \label{eq:closure1}
\end{equation}
where $\eta_k$ refers to the $k$th modal eddy viscosity coefficient.  In our current work, we formulate an RL framework to learn $\eta_k(t)$, and call this approach as multi-modal RL (MMRL). Although the proposed closure problem can be formulated using more traditional adjoint based \cite{cordier2013identification} or sensitivity based approaches \cite{ahmed2020forward}, our chief motivation in this study is to explore the feasiblity of RL workflows for the ROM closure problems. More specifically, in this paper we aim to introduce a variational multiscale RL (VMRL) approach by formulating a new procedure to forge a reward function that does not require access to the training data. Our approach therefore facilitates new RL workflows since RL enhanced computational frameworks might play an integral role in designing many end-to-end data-driven approaches for broader optimization and control problems.      

%\section*{Deep Reinforcement Learning}
%\emph{Deep Reinforcement Learning} --- 
\subsection*{Deep reinforcement learning}
In our context, deep RL presents a modular computational framework to learn $\eta_k(t)$ in Eq.~\ref{eq:closure1}. Here, we briefly describe the formulation of the RL problem and the proximal policy optimization (PPO) algorithm \cite{schulman2017proximal}. In RL, at each time step $t$, the agent observes some representation of the state of the system, $s_t \in \mathcal{S}$, and based on this observation selects an action, $a_t \in \mathcal{A}$. The agent receives the reward, $r_t \in \mathcal{R}$ as a consequence of the action and the environment enters in a new state $s_{t+1}$. Therefore, the interaction of an agent with the environment gives rise to a trajectory as follows
\begin{equation}
    \tau = \{ s_0,a_0,r_0,s_1,a_1,r_1,\dots \}.
\end{equation} 
The goal of the RL is to find an optimal strategy for the agent that will maximize the expected discounted reward over the trajectory $\tau$ and can be written mathematically as follows
\begin{equation}
    \mathfrak{R}(\tau) = \sum_{t=0}^T \gamma^t r_t, \label{eq:reward}
\end{equation}
where $\gamma$ is a parameter called discount rate that lies between $[0,1]$, and $T$ is the horizon of the trajectory. The discount rate determines how much weightage to be assigned to the long-term reward compared to an immediate reward.

In RL, the agent's decision making strategy is characterized by a policy $\pi(s,a) \in {\Pi}$. The RL agent is trained to find a policy to optimize the expected return when starting in the state $s$ at time step $t$ and is called as state-value function. We can write the state-value function as follows
\begin{equation}
    V^{\pi}(s) = \mathbb{E}_\pi \left[ \sum_{k=0}^\infty \gamma^k r_{t+k} | s_t=s, \pi \right].
\end{equation}
Similarly, the expected return starting in a state $s$, taking an action $a$, and thereafter following a policy $\pi$ is called as the action-value function and can be written as 
\begin{equation}
    Q^{\pi}(s,a) = \mathbb{E}_\pi \left[ \sum_{k=0}^\infty \gamma^k r_{t+k} | s_t=s, a_t=a, \pi \right].
\end{equation}
We also define an advantage function that can be considered as an another version of action-value function with lower variance by taking the state-value function as the baseline. The advantage function can be written as 
\begin{equation}
    A^{\pi}(s,a) = Q^\pi(s,a) - V^\pi(s).
\end{equation}

We use $\pi_w(\cdot)$ to denote that the policy is parameterized by $w \in \mathbb{R}^d$ (i.e., the weights and biases of the neural network in deep RL). The agent is trained with an objective function defined as \cite{sutton2018reinforcement}
\begin{equation}
    J(w)~ \dot{=} ~ V^{\pi_w}(s_0), \label{eq:obj_function}
\end{equation}
where an episode starts in some particular state $s_0$, and $V^{\pi_w}$ is the value function for the policy $\pi_w$. The policy parameters $w$ are updated by estimating the gradient of an objective function and plugging it into a gradient ascent algorithm as follows
\begin{equation}
    w \leftarrow w + \beta \nabla_w J(w),    \label{eq:gradient_ascent}
\end{equation}
where $\beta$ is the learning rate of the optimization algorithm. The gradient of an objective function can be computed using the policy gradient theorem \cite{sutton2000policy} as follows
\begin{eqnarray}
    \nabla_w V^{\pi_w}(s_0) = \mathbb{E}_{\pi_w}\big[ \nabla_w \big(\log ~\pi_w(s,a) \big) Q^{\pi_w}(s,a)]. \label{eq:gradient}
\end{eqnarray}
The accurate calculation of empirical expectation in Eq.~\ref{eq:gradient} requires large number of samples. Furthermore, the performance of policy gradient methods is highly sensitive to the learning rate leading to difficulty in obtaining stable and steady improvement. The PPO algorithm introduces a clipped surrogate objective function \cite{schulman2017proximal} to avoid excessive update in policy parameters in a simplified way as follows
\begin{eqnarray}
    J^{{\rm clip}}(w) = \mathbb{E}\big[ {\rm min}(r_t(w) {A}^{\pi_w}(s,a), {\rm clip} (r_t(w),1-\epsilon,1+\epsilon) {A}^{{\pi_w}}(s,a) ) \big], \label{eq:ppo_clip}
\end{eqnarray}
where $r_t(w)$ denotes the probability ratio between new and old policies as given below
\begin{equation}
    r_t(w) = \frac{\pi_{w+\Delta w}(s,a)}{\pi_{w}(s,a)}.
\end{equation}
The $\epsilon$ in Eq.~\ref{eq:ppo_clip} is a hyperparameter that controls how much new policy can deviate from the old. This is done through a function ${\rm clip}(r_t(w),1-\epsilon,1+\epsilon)$ that enforces the ratio between new and old policy ($r_t(w)$) to stay between the limit $[1-\epsilon,1+\epsilon]$. 

\begin{figure}[t]
\centering
\includegraphics[width=0.8\linewidth]{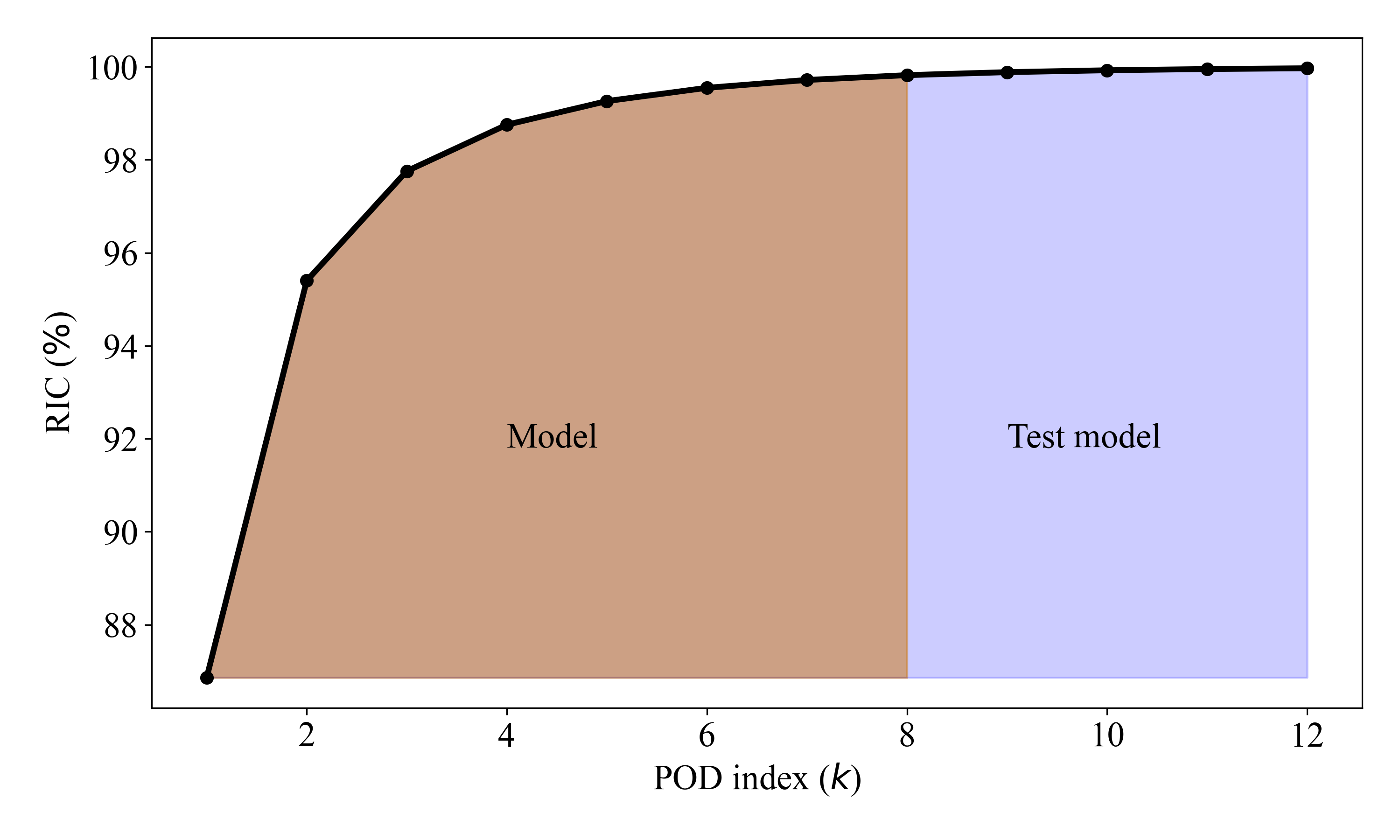}
\caption{Relative information content (RIC) values as a function of the POD index. Test scales,which are used to build our reward function, represent the contribution of the under-resolved ROM scales.}
\label{fig:ric}
\end{figure}

\subsection*{Variational multiscale approach}
Here we present a two-scale variational multiscale formulation as depicted in Figure~\ref{fig:ric}, which utilize two orthogonal spaces, $\boldsymbol X_A$ and $\boldsymbol X_B$. Since the POD basis is orthonormal by construction, we can build these two orthogonal spaces in a natural, straightforward way: $\boldsymbol X_A := \text{span}  \{ \psi_1, \psi_2, \ldots, \psi_{R} \}$, which represents the resolved ROM scales, and 
$\boldsymbol X_B := \text{span}  \{ \psi_{R+1}, \psi_{R+2},\ldots, \psi_{\tilde{R}} \}$, which represents the test scales (i.e., unresolved ROM scales). Following Equation~\ref{eq:rom}, next we use the ROM approximation of $u$ in the space $\boldsymbol X_A \oplus \boldsymbol X_B$,
\begin{align}\label{eq:test}
u(x,t) =\sum_{k=1}^{R} \alpha_k(t) \psi_k(x) + \sum_{k=R+1}^{\tilde{R}} \alpha_k(t) \psi_k(x),
\end{align}
where the first term in the right hand side of Equation~\ref{eq:test} represents the resolved ROM components of $u$, and the second term represents the unresolved test scales. Plugging the ROM approximation of $u$ in Equation~\ref{eq:burgers}, projecting it onto $\boldsymbol X_A$, and using ROM basis orthogonality, we obtain 
\begin{equation} \label{eq:rom2}
  \frac{d \alpha_k}{d t} =  \sum_{i=1}^{\tilde{R}} \mathfrak{L}^{i}_{k}\alpha_{i} + \sum_{i=1}^{\tilde{R}}\sum_{j=1}^{\tilde{R}} \mathfrak{N}^{ij}_{k}\alpha_{i}\alpha_{j}, \quad \forall \ k = 1, \ldots, R.
\end{equation}
In summary, our RL environment consists of three model definition for the evolution of the state variables $\alpha_1, \alpha_2, \ldots, \alpha_R $: (i) base ROM given by Equation~\ref{eq:rom}, (ii) improved ROM by the closure model given by Equation~\ref{eq:rom1}, and (iii) test model given by Equation~\ref{eq:rom2}.  Our key hypothesis relies on the fact that the proposed closure model is accurate and representative if the difference between states obtained by Equation~\ref{eq:rom1} and Equation~\ref{eq:rom2} is minimized. Therefore, the reward function in our RL framework can now be easily constructed by exploiting the difference between these resolved and test scale modal coefficients. More precisely, let's define the following states at time $t$: $s_t^{base} := \{\alpha_1, \alpha_1, \ldots, \alpha_R\}$ as the solution of Equation~\ref{eq:rom}, $s_t^{ROM} := \{\alpha_1, \alpha_1, \ldots, \alpha_R\}$ as the solution of our ROM given by Equation~\ref{eq:rom1}, and $s_t^{test} := \{\alpha_1, \alpha_1, \ldots, \alpha_R\}$ as the solution of Equation~\ref{eq:rom2}. Then we can reward our closure policy according the following definition of the binary reward function:
\begin{align}\label{eq:reward_vms}
    r_t= 
\begin{cases} 
    +10 ,       & \text{if } \sigma ||s_t^{base} - s_t^{ROM}|| < ||s_t^{base} - s_t^{test}|| \\
    -10,        & \text{otherwise}
\end{cases}
\end{align}
where $\sigma > 1$ is a scaling factor that can be chosen between 1 and 2 in practice. In our calculations, we set $\sigma=1.6$. We note that this binary definition of reward function eliminates the need for access to the snapshot data as we will detail further in our results section. The selection of $\pm10$ in our binary definition is arbitrary since the RL workflows are designed to maximize the sum of the reward over each episodic experiment.  

\begin{figure}[htbp!]
\centering
\includegraphics[width=0.98\linewidth]{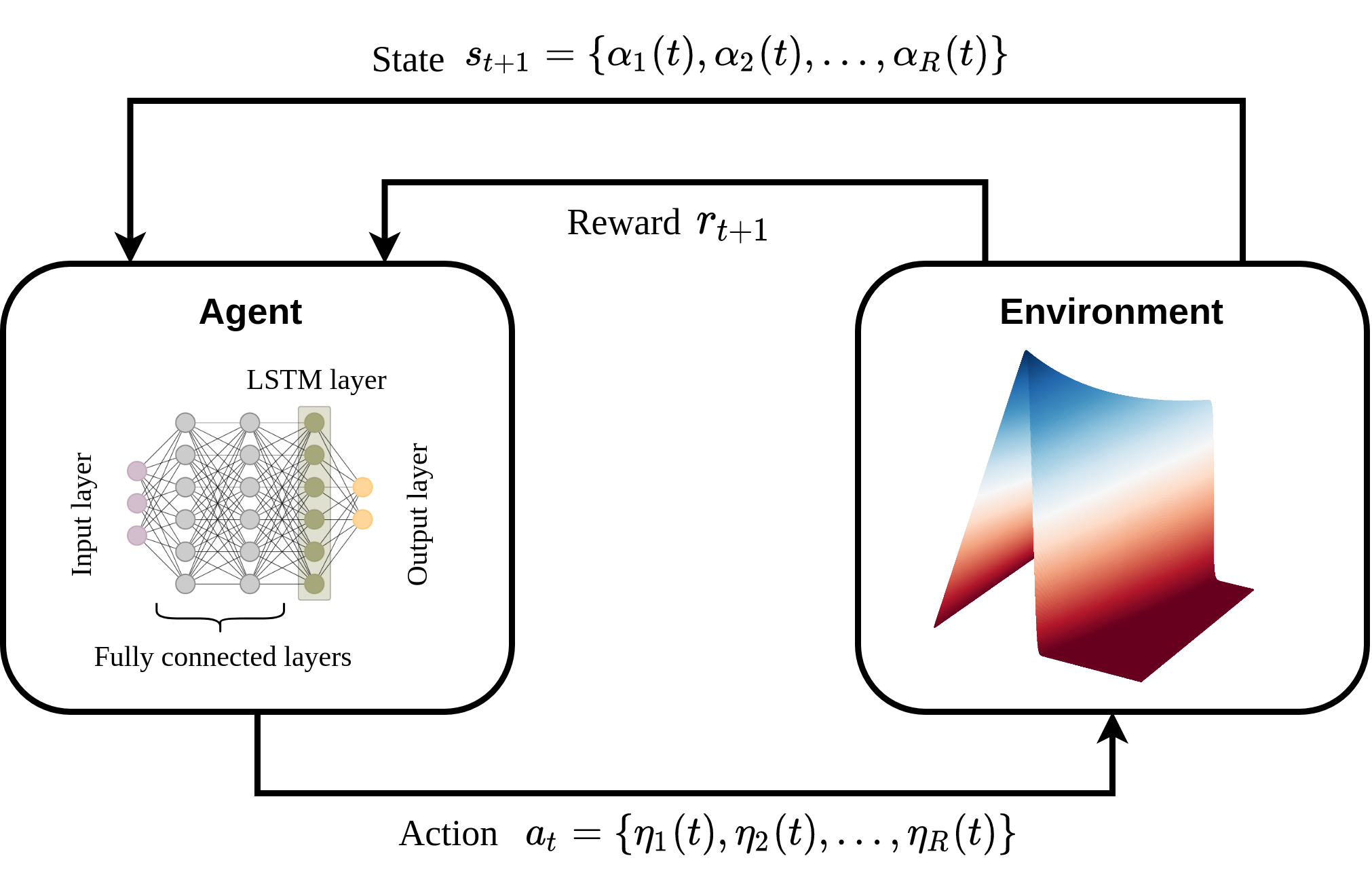}
\caption{Schematic of the deep RL framework for MMRL ROM closure approach. The RL agent observes modal coefficients and selects model eddy viscosity coefficient as an action. %The neural network architecture is only for demonstration purpose only.
}
\label{fig:rl_closure}
\end{figure}

\begin{figure*}[htbp!]
\centering
\includegraphics[width=0.95\linewidth]{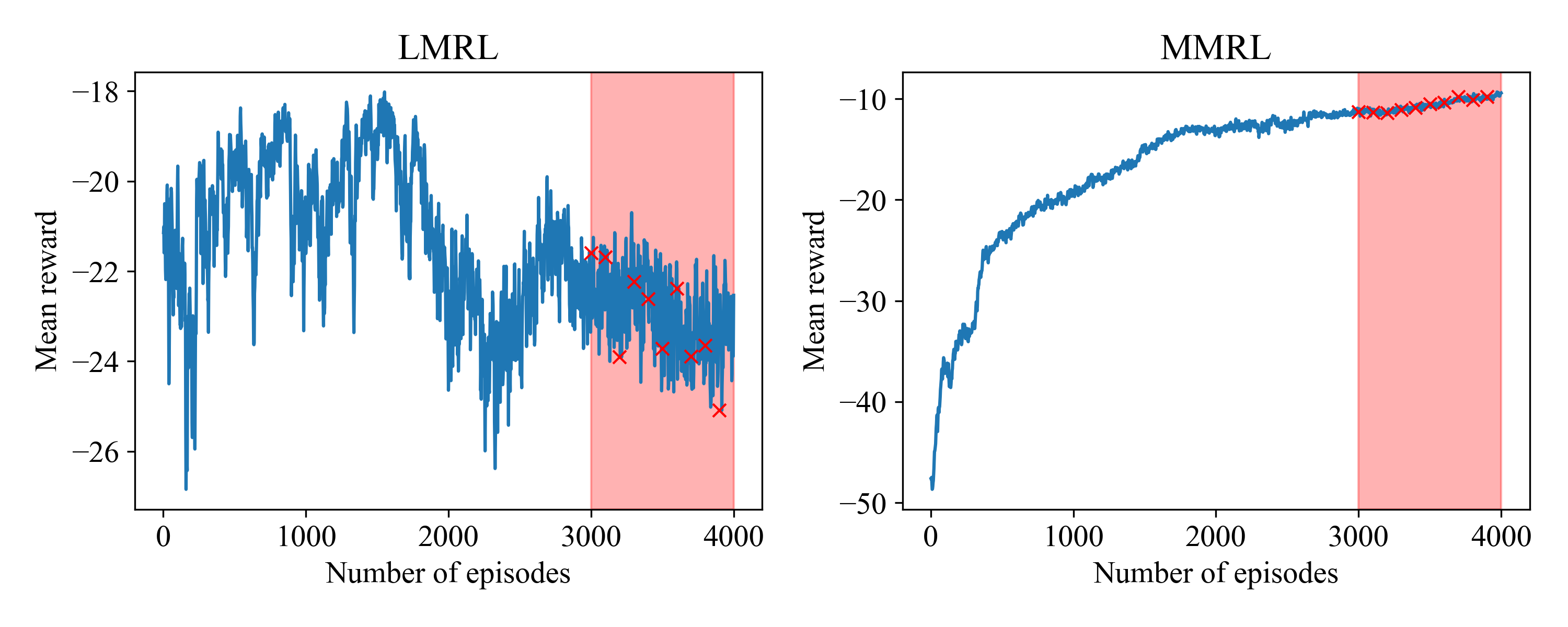}
\caption{Evolution of the moving average mean reward for LMRL (left) and MMRL (right) approaches. The models used for testing are indicated by red symbols.}
\label{fig:reward}
\end{figure*}

\begin{figure*}[htbp!]
\centering
\includegraphics[width=0.9\linewidth]{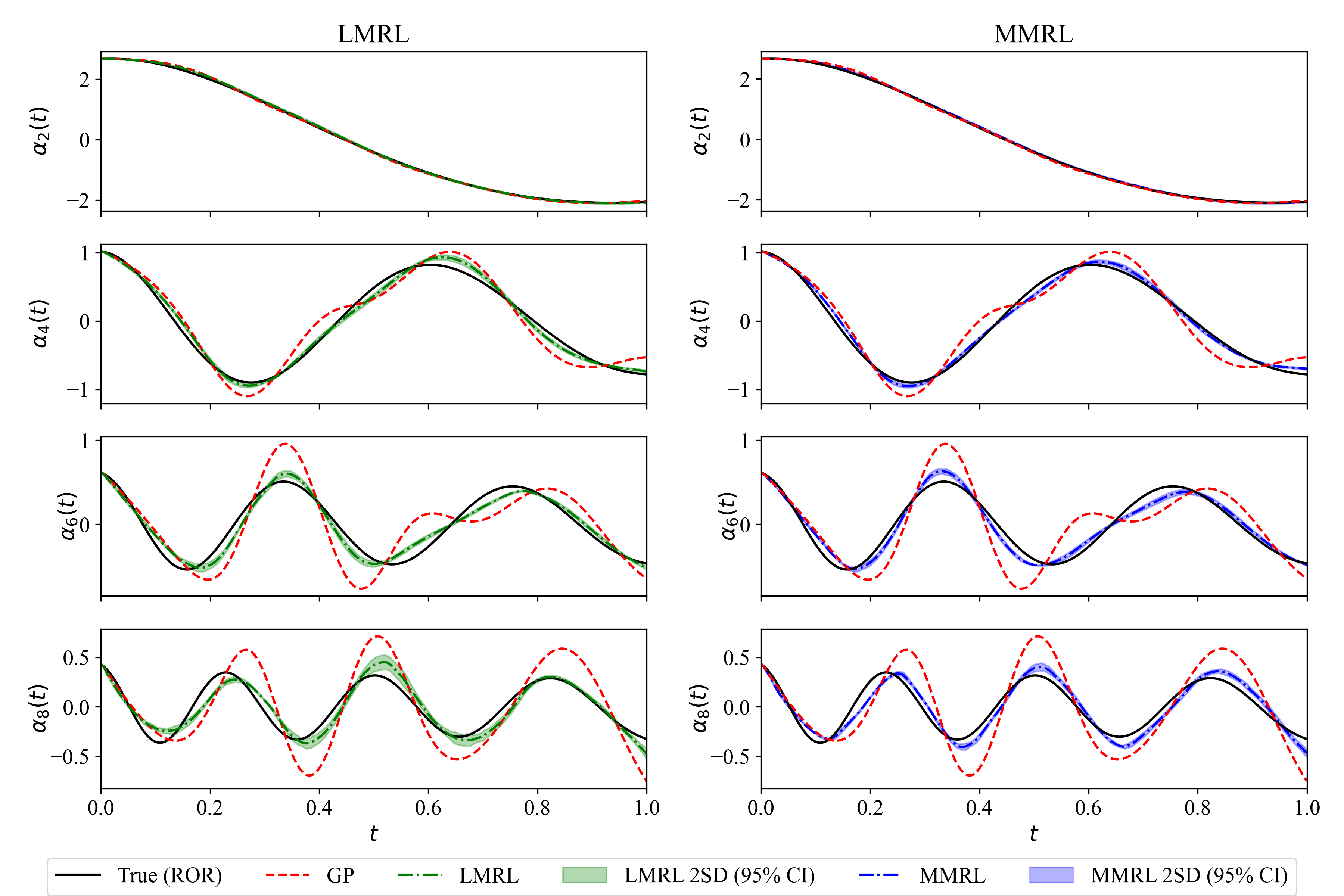}
\caption{Evolution of the second, fourth, sixth and the last POD modal coefficients at $Re = 1500$ for LMRL (left) and MMRL (right) approaches.}
\label{fig:modes}
\end{figure*}

\begin{figure*}[htbp!]
\centering
\includegraphics[width=0.5\linewidth]{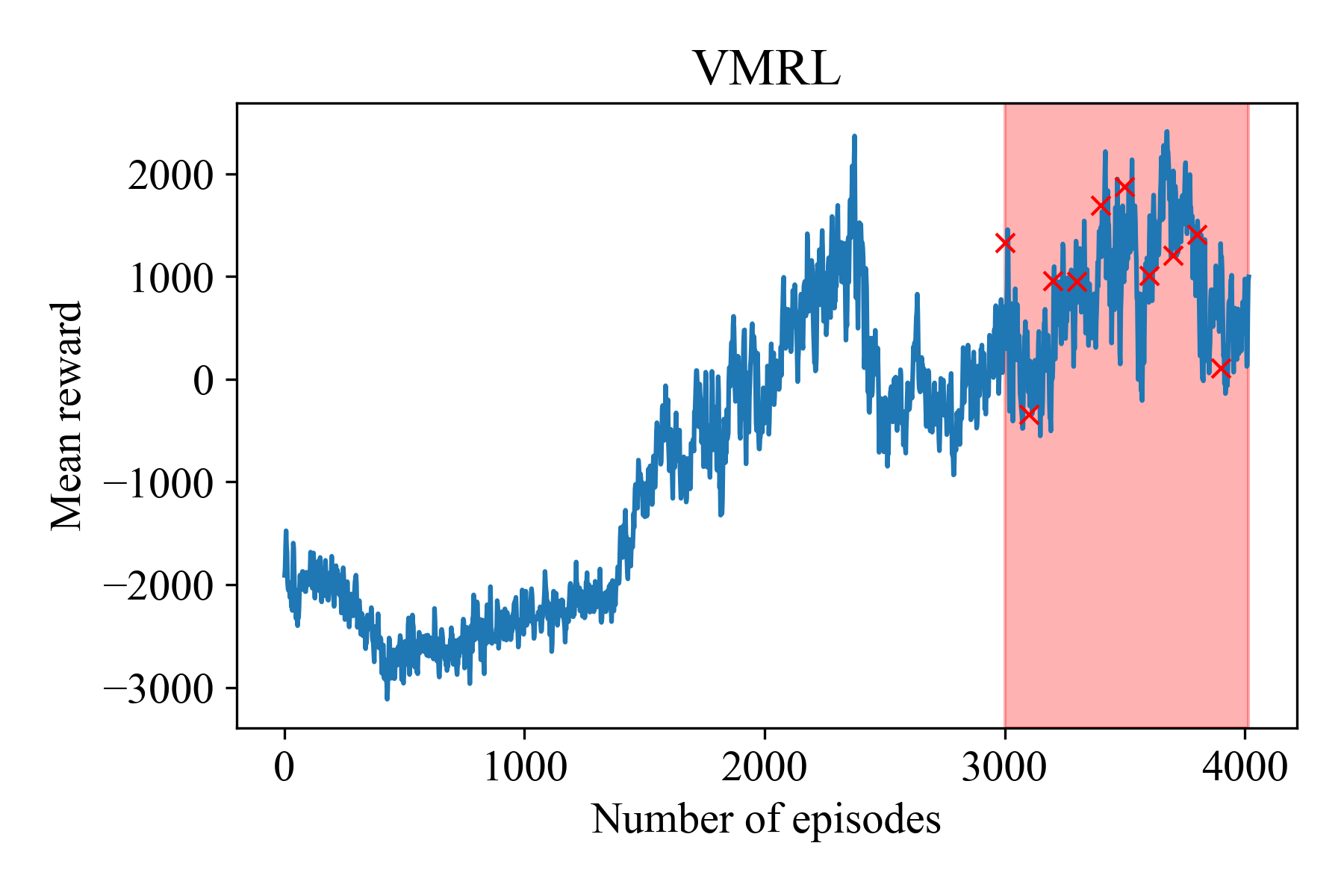}
\caption{Evolution of the moving average mean reward for VMRL approaches. The models used for testing are indicated by red symbols.}
\label{fig:reward2}
\end{figure*}

\begin{figure*}[htbp!]
\centering
\includegraphics[width=0.9\linewidth]{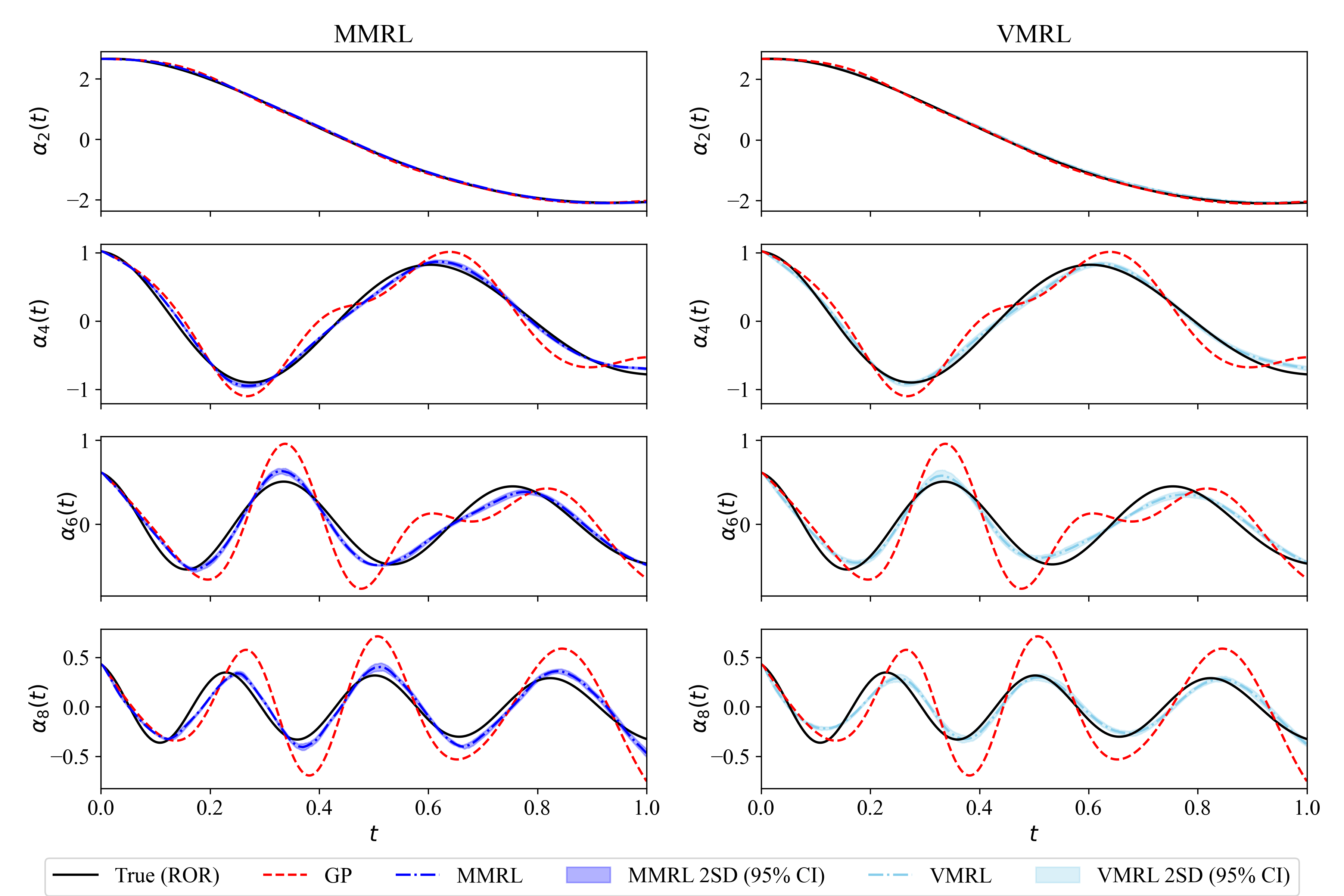}
\caption{Evolution of the second, fourth, sixth and the last POD modal coefficients at $Re = 1500$ for MMRL (left) and VMRL (right) approaches.}
\label{fig:modes2}
\end{figure*}

\begin{figure}[htbp!]
\centering
\includegraphics[width=1.0\linewidth]{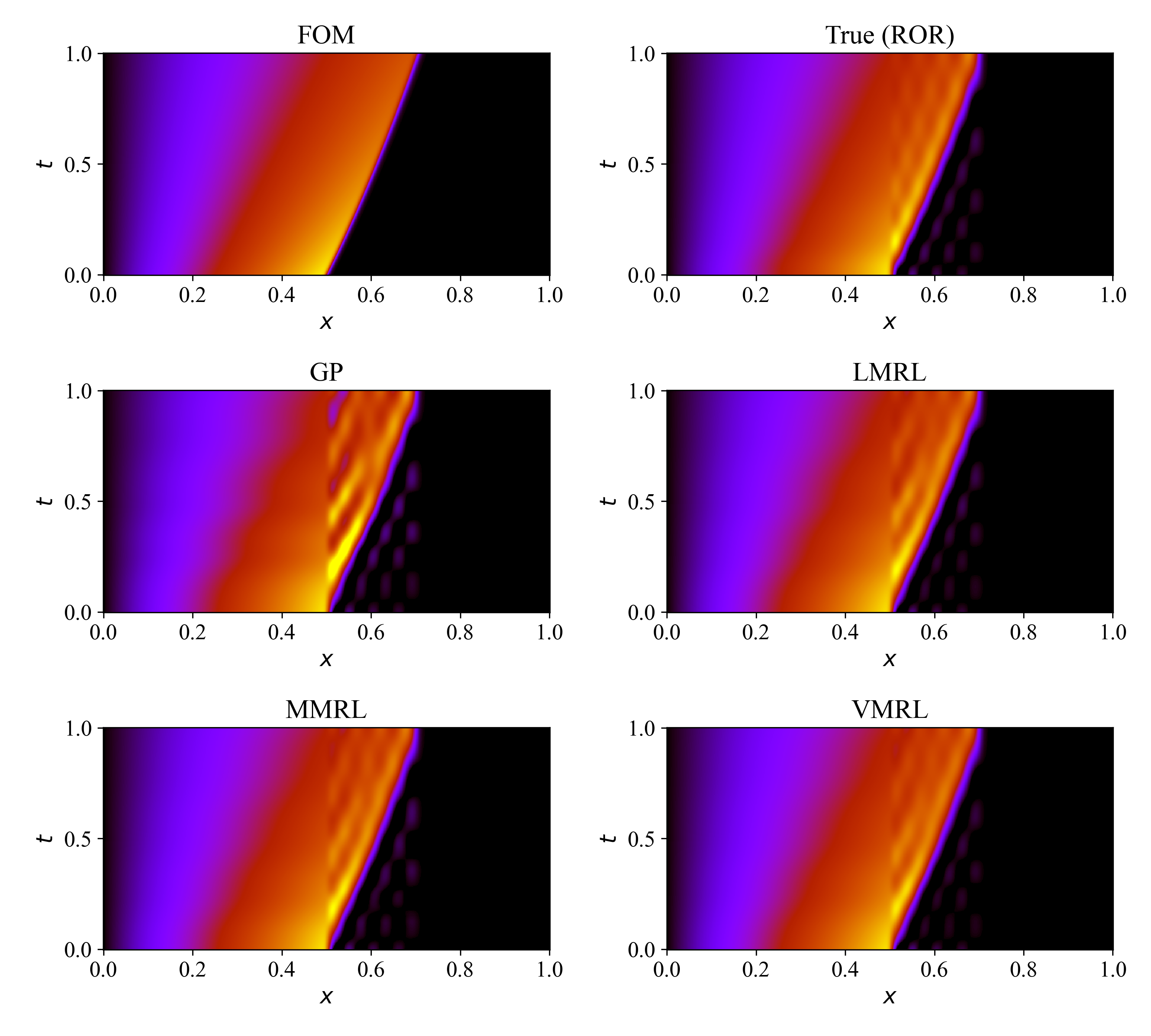}
\caption{Spatiotemporal visualization of the velocity field at $Re = 1500$ for different modeling approaches.}
\label{fig:field}
\end{figure}

\section*{Results}

% \section*{Results} \label{sec:results}
%\emph{Results} --- 
%\subsection*{Implementation of Deep RL}
Figure~\ref{fig:rl_closure} displays the complete deep RL framework for the MMRL approach where the agent observes the POD modal coefficients as the state of the system and takes the action of selecting modal eddy viscosity coefficients. Due to the modal truncation, the effect of unresolved scales on resolved scales is not captured and there is a discrepancy between the true modal coefficients and the ROM modal coefficients. The goal of the agent is to minimize this difference, and therefore, the $l_2$ norm of the deviation between true and ROM modal coefficients is used as the reward function. Since we are maximizing the reward, the negative of the $l_2$ norm is first assigned as a reward at each time step that is given by
\begin{align}\label{eq:reward}
r_t = - ||s_t^{ROM} - (\hat{u},\psi_k)||
\end{align}
where $\hat{u} :={u(x,t)}$ refers to snapshot data at time step $t$. The choice of the reward function can have a significant effect on the performance of the agent \cite{novati2021automating} and needs to be carefully designed depending upon the problem. The performance of the MMRL approach is compared against the linear-modal RL (LMRL) approach where the agent selects only a scalar value of eddy viscosity amplitude as an action and linear viscosity kernel \cite{san2014proper} is utilized to assign the modal eddy viscosity coefficients. Specifically, in LMRL approach we have $\eta_k(t)=\eta_e(t)(k/R)$, where $\eta_e$ is the eddy viscosity amplitude selected by the agent as an action.

%\subsection*{Numerical results}
In Fig.~\ref{fig:reward}, the trajectory of the mean reward is shown for training an RL agent with the LMRL and MMRL approaches. The agent is trained for Reynolds number $Re = 1000$. It can be seen that the maximum reward attained with the MMRL approach is almost twice the magnitude of the reward achieved with the LMRL approach. Figure~\ref{fig:modes} depicts the evolution of selected POD modal coefficients for $Re = 1500$. The prediction from both LMRL and MMRL approaches is in better agreement with the true projection modal coefficients compared to GP with the prediction from MMRL being more accurate compared to LMRL. However, we highlight that both LMRL and MMRL approaches utilize the reward function given by Equation~\ref{eq:reward}, which requires access to the true snapshot data. We highlight that in our evaluation, both mean and two-standard deviation (i.e., 95\% confidence interval) of 10 different RL models are shown (e.g., see red symbols in Fig.~\ref{fig:reward} for those models).

On the other hand, Fig.~\ref{fig:reward2} illustrates the trajectory of the mean reward for training an RL agent with the VMRL approach that utilizes a binary reward function given by Equation~\ref{eq:reward_vms}.  Figure~\ref{fig:modes2} shows a comparison between MMRL and VMRL approaches at $Re = 1500$. We highlight that both approaches utilizes multi-modal action space (i.e., discovering $\eta_k(t)$ for $k=1,2, \ldots, R$). Figure~\ref{fig:modes2} clearly demonstrates that the VMRL approach obtains an accurate policy without requiring access to the true labeled data in defining the reward function. This key aspect of the proposed VMRL approach paves the way of designing novel RL workflows exploiting the modal interaction between resolved and test scales.   

The spatiotemporal velocity field with different ROMs is shown in Fig.~\ref{fig:field}. It should be noted that we can at the most recover the true projection of the full order model (FOM) solution. This true reduced order representation (ROR) is also shown in Fig.~\ref{fig:field}. In our analogy given in Fig.~\ref{fig:closure}, the blue curve represents the ROR, the red curve represents the GP model. For quantitative analysis, the root mean squared error (RMSE) for different ROM approaches at different Reynolds numbers (different from the training setting at $Re=1000$) is reported in Table~\ref{table:rmse}. The RMSE for LMRL, MMRL, and VMRL approaches is significantly smaller compared the the GP model. We also observe that the VMRL approach provides marginally more accurate solution than the LMRL and MMRL approaches at higher Reynolds number.

\begin{table}[htbp!]
\renewcommand{\arraystretch}{1.25}
\caption{$\ell_2$ norm for the deviation of the velocity with respect to its true projection value for $t\in [0,1]$. We note that both snapshot data generation for POD analysis and RL training are performed at $Re=1000$. Here LMRL and MMRL models use the reward function defined in Equation~\ref{eq:reward} utilizing the true snapshot data, whereas the VMRL model uses the reward function defined in Equation~\ref{eq:reward_vms} that utilizes the variational multiscale formalism.}
\centering
\begin{tabular}{p{0.1\textwidth} p{0.25\textwidth} l l l}  
\hline
ROM  & Action & RMSE ($Re = 1200$) & RMSE ($Re = 1500$) & RMSE ($Re = 2000$) \\
\hline
GP   & - &  $11.837\times10^{-3}$ & $16.217\times10^{-3}$& $22.809\times10^{-3}$\\ 
LMRL  & $a_t \in \{\eta_e(t)\}$ & $4.645\times10^{-3}$  & $6.144\times10^{-3}$ & $10.092\times10^{-3}$\\ 
MMRL  & $a_t \in \{\eta_1(t), \eta_2(t),..., \eta_R(t)\}$ &  $3.111\times10^{-3}$  & $5.258\times10^{-3}$ & $9.746\times10^{-3}$ \\ 
VMRL  & $a_t \in \{\eta_1(t), \eta_2(t),..., \eta_R(t)\}$ & $5.341\times10^{-3}$  & $5.262\times10^{-3}$ & $8.063\times10^{-3}$ \\ 
\hline
\end{tabular}
\label{table:rmse}
\end{table}

\section*{Discussion}
%\emph{Conclusion} --- 
This study introduces scale-aware reinforcement learning (RL) framework to automate the discovery of closure models in projection based ROMs. We treat the closure as a control input in the latent space of the ROM and build the parameterized model with a dissipation term.
The feasibility of the RL framework is first demonstrated with linear-modal RL (LMRL) where a linear eddy viscosity constraint is utilized for parameterization and with multi-modal RL (MMRL) which finds mode dependant eddy viscosity model coefficient. The agent is incorporated in a reduced-order solver, observes the POD modal coefficients, and accordingly computes the closure term. Here, both RL approaches minimize the discrepancy between the true POD modal coefficients and prediction from closure ROM, and the obtained closure model generalizes to different Reynolds number. We then demonstrate how to formulate an RL framework without requiring access to the true data using the variational multiscale formalism. We find that this variational-multiscale RL (VMRL) is a robust closure discovery framework that utilizes a reward function based on the modal energy transfer effect. Building on the promising results presented in this study to develop ROM closure models using deep RL, our future work will concentrate on incorporating uncertainties associated with observations into account while selecting the action of an agent.

\bibliography{sample}

\section*{Acknowledgements}

%This material is based upon work supported by the U.S. Department of Energy, Office of Science, Office of Advanced Scientific Computing Research under Award Number DE-SC0019290.
%O.S. gratefully acknowledges their Early Career Research Program support.
%O.S. gratefully acknowledges their support.
O.S. gratefully acknowledges the financial support of the National Science Foundation under Award Number DMS-2012255 and the Early Career Research Program (ECRP) support of the U.S. Department of Energy under Award Number DE-SC0019290.

% \section*{Author contributions statement}
% O.S. conceived the proposed variational multiscale reinforcement learning closure modeling approach. O.S. and S.P conducted the numerical experiments, analysed the data, and wrote the initial draft of the manuscript.  A.R. provided feedback and contributed in shaping the research, analysis and manuscript. All coauthors reviewed the manuscript, discussed the results, and contributed to the final manuscript.

%O.S. and S.P conducted the numerical experiments, analysed data, and wrote the initial draft of the manuscript.
%S.L. and J.M.L conceived the research. S.L. wrote the initial draft of the manuscript. S.E.A performed simulations and analysed the data. O.S. supervised study, provided feedback and contributed in shaping the research, analysis and manuscript. All coauthors reviewed the manuscript, discussed the results, and contributed to the final manuscript.

\section*{Competing interests} 
The authors declares no competing interests.

\section*{Additional information}

The data that supports the findings of this study are available within the article. The datasets used and/or analysed during the current study are available from the corresponding author on reasonable request.

%The open-source implementation and datasets generated and/or analysed during the current study are available in the GitHub repository, \url{https://github.com/omersan}.

%The open-source implementation is available at \url{https://github.com/Shady-Ahmed}.

%The datasets used and/or analysed during the current study available from the corresponding author on reasonable request.
%Correspondence and requests for materials should be addressed to O.S. 

% To include, in this order: \textbf{Accession codes} (where applicable); \textbf{Competing interests} (mandatory statement). 

% The corresponding author is responsible for submitting a \href{http://www.nature.com/srep/policies/index.html#competing}{competing interests statement} on behalf of all authors of the paper. This statement must be included in the submitted article file.

% % \begin{figure}[ht]
% % \centering
% % \includegraphics[width=\linewidth]{stream}
% % \caption{Legend (350 words max). Example legend text.}
% % \label{fig:stream}
% % \end{figure}

% \begin{table}[ht]
% \centering
% \begin{tabular}{|l|l|l|}
% \hline
% Condition & n & p \\
% \hline
% A & 5 & 0.1 \\
% \hline
% B & 10 & 0.01 \\
% \hline
% \end{tabular}
% \caption{\label{tab:example}Legend (350 words max). Example legend text.}
% \end{table}

% Figures and tables can be referenced in LaTeX using the ref command, e.g. Figure \ref{fig:stream} and Table \ref{tab:example}.

\end{document}